\documentclass[letterpaper, 10 pt, conference]{ieeeconf}
\usepackage[utf8]{inputenc}
\UseRawInputEncoding
\IEEEoverridecommandlockouts    %used for \thanks
\usepackage{cite}
\usepackage{amsmath,amssymb,amsfonts,amstext}
\usepackage{graphicx}
\usepackage{textcomp}
\usepackage[utf8]{inputenc}
\usepackage{babel}
\usepackage{balance}
\usepackage{comment}
\usepackage[position=bottom]{subfig}

\usepackage{mathtools}
\usepackage{float}

\usepackage{amsthm}

\newtheorem{assumption}{Assumption}

\usepackage{algorithm}
\usepackage{algorithmicx} 
\usepackage{algpseudocode} %

\DeclareMathAlphabet{\mathbbmsl}{U}{bbm}{m}{sl}
\usepackage{todonotes}

\makeatletter
\let\NAT@parse\undefined
\makeatother
\usepackage{hyperref}
\hypersetup{
  colorlinks=true,
    linkcolor= blue,
    allcolors=blue,
    citecolor = blue,
    filecolor=black,      
    urlcolor=blue,
    }

\title{\bf AI Recommendation Systems for Lane-Changing\\Using Adherence-Aware Reinforcement Learning}

\author{Weihao Sun$^1$, Heeseung Bang$^2$, \textit{Member, IEEE}, and Andreas A. Malikopoulos$^{1,2}$, \textit{Senior Member, IEEE}
    \thanks{This research was supported by in part by NSF under Grants CNS-2401007, CMMI-2348381, IIS-2415478 and in part by MathWorks.}
    \thanks{$^1$ Weihao Sun and Andreas Malikopoulos are with the Systems Engineering Program, Cornell University, Ithaca, NY 14850 USA. {\tt\small email: \{ws493,amaliko\}@cornell.edu}}
	\thanks{$^2$ Heeseung Bang and Andreas Malikopoulos are with the School of Civil and Environmental Engineering, Cornell University, Ithaca, NY 14850, USA. {\tt\small email: \{h.bang,amaliko\}@cornell.edu}}
}

\date{May 2025}

\begin{document}

\maketitle
\begin{abstract}
In this paper, we present an adherence-aware reinforcement learning (RL) approach aimed at seeking optimal lane-changing recommendations within a semi-autonomous driving environment to enhance a single vehicle’s travel efficiency. The problem is framed within a Markov decision process setting and is addressed through an adherence-aware deep Q network, which takes into account the partial compliance of human drivers with the recommended actions. This approach is evaluated within CARLA’s driving environment under realistic scenarios. 
\end{abstract}

\section{Introduction} \label{sec:introduction}

Autonomous driving is classified into five levels, ranging from Level 1 to Level 5 \cite{autonomous_classification}. However, complete Level 5 automation has yet to be realized. Current research and development primarily focus on partially autonomous vehicles at Levels 2 to 4, where human drivers retain varying degrees of responsibility. In such a semi-autonomous driving environment, an artificial intelligence (AI) system usually provides recommendations on driving decisions. In recent years, researchers have explored various control methods for mixed driving environments \cite{intro_semi_env, le2024stochastic, bang2024confidence} where autonomous vehicles and human driving vehicles coexist. In such a mixed traffic environment, the ability of AI to make rapid and appropriate decisions is critical to ensure safety and efficiency from the perspective of human drivers. Effective improvement on AI recommendations relies on adaptive decision-making that accounts for human behaviors \cite{dave_bang_2024}. 

In the context of AI assistance in driving, lane-changing decisions exemplify the complexity of dynamic human-AI interaction, which directly affects travel safety and efficiency, making such a problem a critical focus area in intelligent transportation. Traditional optimization-driven approaches often lack addressing variability in human behaviors, leading to immature outcomes that are only partly accepted by human drivers. In recent years, reinforcement learning (RL) methods have emerged as a potential solution due to their capacity to find optimal decision policies through a variety of complex environments. However, conventional RL approaches barely model human adherence or treat them as system noise \cite{faros_q-learning_2023}.

Many studies have applied machine learning techniques to address decision-making problems in the transportation area, driven by advances in computational power and improved collected data quality in recent years. Some existing literature has explored the use of conventional RL techniques to learn optimal policies for lane-changing decisions. For instance, some studies utilized a deep Q network-based learning strategy to optimize a vehicle's velocity while maintaining safety \cite{max_v_itsc2018,two_agents_2itsc2018}. Another study explored deep RL approaches to lane-changing problems using rule-based constraints \cite{Wang_2019}.
The overall objectives of these efforts are similar to those of our paper. However, scenarios in which human drivers may not always comply with the movements have not been considered. Another study proposed a harmonious lane-changing strategy to address a single vehicle's limited sensing information in an autonomous driving environment \cite{harmonious_5tits2022}. In another study, researchers trained a deep RL agent to plan lane-changing trajectories \cite{trajectory_4tits2022} or to make merging decisions \cite{Sumanth2021}, which are different cases from the perspective of problem modeling. On the other hand, some research efforts suggested learning the human internal state affecting their decision 
 \cite{dave_bang_2024}. Other studies attempted to collect information to learn such internal states \cite{information_gather}, and some researchers \cite{Nishanth2023AISmerging} utilized approximate information state models to learn human driver behaviors in highway merging problems. Moreover, some efforts utilize reinforcement learning (RL) approaches to offer personalized lane change decisions for human drivers \cite{li2023personalized}, but the work we report in this paper differs in its focus on adopting an algorithm that learns the compliance levels of human drivers.

In this paper, we propose a lane-changing decision recommendation system using adherence-aware RL. In typical RL, the solution may not be optimal if the human driver does not comply with the recommendation. Thus, we consider this partial compliance case to provide a better solution.
We first model the lane-changing scenario as a Markov decision process (MDP) and incorporate human drivers' compliance patterns within the recommendation process. 
We then train a deep Q network (DQN) to learn an optimal recommendation policy that aligns with the likelihood of human driver compliance during online learning. The adherence-aware Q learning algorithm in the DQN is adopted from \cite{faros_q-learning_2023} to model the level of human adherence to lane-changing recommendations. We conduct comprehensive learning and testing using CARLA simulator \cite{carla} to demonstrate the effectiveness of our method.
% In this paper, we address these challenges by modeling the lane-changing decision task as a Markov decision process (MDP),  and incorporating human driver compliance within the recommendation process. 

Our main contributions include 1) the development of an adherence-aware DQN-based RL framework that learns optimal lane-changing recommendation policies while accounting for human drivers' compliance levels and 2) empirical demonstration of improvements compared to regular RL and human drivers' baseline actions. 
Our framework introduces an approach to improve the AI recommendations system to make lane-changing decisions more aligned with the human driver's experience.
% \hb{Talk more about impact, how it improves human's driving experience.}

% \begin{enumerate}
%     \item We propose an adherence-aware DQN-based RL method to learn the optimal policy for lane-changing high-level decisions regarding human driver's partial adherence towards recommendation action.
%     \item We use the CARLA environment to perform simulation on both the traffic environment and human driver's behaviors.
% \end{enumerate}

% \newpage

\section{Problem Formulation}

Lane-changing scenarios involve uncertainty due to various human driver behaviors and the complexities of traffic environments. 
To address this, we model the lane-changing problem as an MDP, considering human drivers' partial adherence.
The MDP is described by a set of system states, $\mathcal{X}$, a behavior set $\mathcal{U}$, a reward function $r: \mathcal{X\times\mathcal{U}\to\mathbb{R}}$, and a transition probability function $p:\mathcal{X}\times\mathcal{U}\times\mathcal{X}\to\Delta(\mathcal{X})$. 
We denote the set of lanes by $\mathcal{L}=\{1,\dots,L\}$, $L\in\mathbb{N}$. 
The states describe the spatial and temporal aspects of both the ego vehicle and the surrounding vehicles. The state of the ego vehicle is denoted as a function of time, i.e.,
\begin{equation}
    x^\mathrm{e}(t) = [s^\mathrm{e}(t), v^\mathrm{e}(t), l^\mathrm{e}(t)]^T,
\end{equation}
where $s^\mathrm{e}(t) \in \mathbb{R}_{>0}$ is the longitudinal position, $v^\mathrm{e}(t) \in \mathbb{R}_{>0}$ is the speed, $l^\mathrm{e}(t)\in\mathcal{L}$ is the lane of the ego vehicle at time $t$.
To account for surrounding vehicles that affect lane-changing decisions, we select five key vehicles that have the most significant impact to the ego vehicle, i.e., $\mathcal{V}=\{\mathrm{V}_{\text{ego},l},\mathrm{V}_{\text{left},l}, \mathrm{V}_{\text{left},f}, \mathrm{V}_{\text{right},l}, \mathrm{V}_{\text{right}, f}\}$ where each of them refers to the leader of ego vehicle ($\mathrm{V}_{\text{ego},l}$), leader/follower in left lane ($\mathrm{V}_{\text{left},l}/ \mathrm{V}_{\text{left},f}$), leader/follower in right lane ($\mathrm{V}_{\text{right},l}/ \mathrm{V}_{\text{right}, f}$), respectively.
Leader and follower in adjacent lanes are determined by the relative position of the ego vehicle. For example, if the vehicle is ahead of the ego vehicle, we consider it to be a leader in that lane.
The state of each surrounding vehicle $V\in\mathcal{V}$ is denoted by 
\begin{equation}
    x^{\mathrm{V}}(t) = [s^\mathrm{V}(t), v^\mathrm{V}(t), l^\mathrm{V}(t)]^T. 
\end{equation}
%representing the five surround vehicles respectively.
The entire system state, including every vehicle that is considered, is denoted by 
% \begin{equation}
%     x(t) = [x^\mathrm{e}(t), x^{\mathrm{V}_{\text{ego},l}}, x^{\mathrm{V}_{\text{left},l}}, x^{\mathrm{V}_{\text{left},f}}, x^{\mathrm{V}_{\text{right},l}}, x^{\mathrm{V}_{\text{right}, f}}]^T.
% \end{equation}

\begin{equation}
    x(t) = [x^\mathrm{e}(t), x^{\mathrm{V}}(t)]^T.
\end{equation}

Note that, as we consider a recommendation system for lane-changing problems, the human driver has the ability to control the acceleration and steering of the vehicle, whereas our system action, which is a recommendation, only suggests an action of lane changing. Thus, we denote the lane-changing action of the ego vehicle as
\begin{equation}
    u^\mathrm{e}(t) \in \{L,R,K\},
\end{equation}
where each element refers to the action of changing lanes to the left (L) and to the right (R) or keeping on the current lane (K). At time $t$, the recommended action is denoted by $u^r(t)=g^r(x(t))$, and the baseline action is denoted by $u^b(t) = g^b(x(t))$, where $g^r$ and $g^b$ are the recommendation strategy and the human's baseline strategy, respectively.

The reward for each time step, $r_t(x(t), u^\mathrm{e}(t))$, encourages the ego vehicle to maintain the speed close to the desired speed by considering $v^\mathrm{e}(t)$ while considering safety aspects. The function also penalizes any dangerous movements, lane changes that do not lead to a faster flow, or missing the chance of lane changing that could possibly reduce travel time. The reward function is defined as 
\begin{equation}\label{eq:reward}
    \begin{multlined}
    r_t(x(t), u^\mathrm{e}(t)) = - \alpha_1\cdot |v^\mathrm{e}(t) - v^\mathrm{des}(t)| \\ - \alpha_2\cdot C_{\mathrm{lane}}(x(t)) - \alpha_3\cdot C_\mathrm{safe}(x(t))  \\ - \alpha_4\cdot C_\mathrm{missing} (x(t), u^\mathrm{e}(t)),
    \end{multlined}
\end{equation}
where $\alpha_1,\alpha_2, \alpha_3, \alpha_4$ are tunable parameters to balance each reward term in the function; $C_{\mathrm{lane}}(x(t))$,  $C_\mathrm{safe}(x(t))$, and $C_\mathrm{missing} (x(t), u^\mathrm{e}(t))$ are the cost functions for unnecessary changing, unsafe behaviors, and missing changing time, respectively.
Specifically, the function $C_{\mathrm{lane}}(x(t))$ returns a penalty when lane changing is executed (either left or right) without contributing to the ego vehicle's travel time. When $u^\mathrm{e}(t)\in\{L,R\}$:
\begin{equation}
    C_{\mathrm{lane}}(x(t)) = 
    \begin{cases}
        1, & v^\mathrm{e}(t)>v^\mathrm{th}_1, \\ 
           &\text{ or } v^\mathrm{e}(t)-v^\mathrm{ego, l}(t) < \Delta v^\mathrm{min}\\
        0, &\text{otherwise,}
    \end{cases}
\end{equation}
where $v^\mathrm{th}_1$ is the threshold velocity when encountering traffic congestion, $\Delta v^\mathrm{min}$ is the minimum velocity difference between the ego vehicle and leading vehicle for a lane changing without penalizing the reward function. When $u^e(t) = K$, $C_{\mathrm{lane}}(x(t)) = 0$.

The safety cost function, $C_\mathrm{safe}(x(t))$, cumulates the penalties of getting too close to surrounding vehicles.
\begin{equation}
    C_\mathrm{safe}(x^\mathrm{s}(t)) = \sum_{i\in\mathcal{V}}\mathbb{I}(\delta^\mathrm{e}_i<t^\mathrm{th}),
\end{equation}
where $\delta^\mathrm{e}_i$ is the time headway between the ego vehicle and another surrounding vehicle $i$. 
The time headway can be computed by the longitudinal distance divided by the relative speed of the vehicles, i.e., $\delta^\mathrm{e}_i=|\{s^\mathrm{e}(t)-s^i(t)\}/\{v^\mathrm{e}(t)-v^i(t)\}|$.
% It is calculated by $\frac{d}{\Delta v}$, where $d$ is the distance and $\Delta v$ is the relative speed. $t^\mathrm{th}$ is the time threshold.

The reward function also penalizes not changing to a lane with faster vehicle flow if the ego car slowly drives behind a leading vehicle. The cost function for missing changing time is defined as 
\begin{equation}
C_{\text{missing}}(x(t), u^e(t)) =
\begin{cases}
    1, & \text{if } u^e(t) = K,\; v^e(t) < v^\mathrm{th}_2,\; \\
        & \text{and lane change is available} \\
    0, & \text{otherwise,}
\end{cases}
\end{equation}
where $v^\mathrm{th}_2$ denotes the minimum speed among the current lane and two adjacent lanes, i.e., $v^\mathrm{th}_2 = \min(v_L,v_R,v_K)$. The speed at each lane can be computed by the average speed of the leader and follower in the lane. In case either leader or follower does not exist, we consider the speed of the missing vehicle to be the maximum allowed speed.

To consider the partial adherence behavior of the human driver, we denote the compliance level by $\theta\in[0,1]$, representing the probability of human drivers following the recommendation $u^\mathrm{r}(t)$. Otherwise, the human driver takes their underlying driving decision (baseline action) $u^b(t)$ \cite{faros_q-learning_2023, chen2023learning}. The equation to describe if the human driver takes the recommended action can be written as
\begin{equation}
u_t = 
\begin{cases}
    u^r_t=g^r(x(t)) & \text{with probability } \theta ,\\
    u^b_t=g^b(x(t)) & \text{with probability } 1-\theta.
\end{cases}
\end{equation}

In our framework, we impose the following assumptions.
\begin{assumption} \label{asmp:theta}
    The stationary adherence level $\theta$ and the transition probability $P$ are unknown.
\end{assumption}
In real-world driving scenarios, the true adherence behavior of the human driver is unavailable beforehand due to dynamic changes in the traffic environments and human intentions. Assumption \ref{asmp:theta} reflects the practical limitation on the observability of the human driver's state for the learning algorithm. Using a derived Bellman-like equation, the optimal policy can be obtained if the adherence level and system dynamic are known \cite{faros_q-learning_2023}. 
The Q value of the next state is estimated from stored transitions in DQN to address the unknown transition probability.
% By allowing the system to learn both the optimal policy and human driver adherence level, the model is capable of supporting making decisions under various human driver compliance levels.

\begin{assumption}\label{asmp:shared-goal}
    The human driver and recommendation systems are assumed to share an identical long-term objective, which is to optimize the expected total reward defined in  \eqref{eq:reward}.
\end{assumption}
The recommendation systems should provide lane-changing recommendations to assist human drivers in achieving their driving goals. Assumption \ref{asmp:shared-goal} ensures that even if human drivers may not comply with the recommendation, they will not take actions conflicting with achieving the common shared goal, which is to improve travel efficiency in this context. 

\section{Solution Approach}

\begin{figure}[th]
    \centering
    \includegraphics[width=0.7\linewidth]{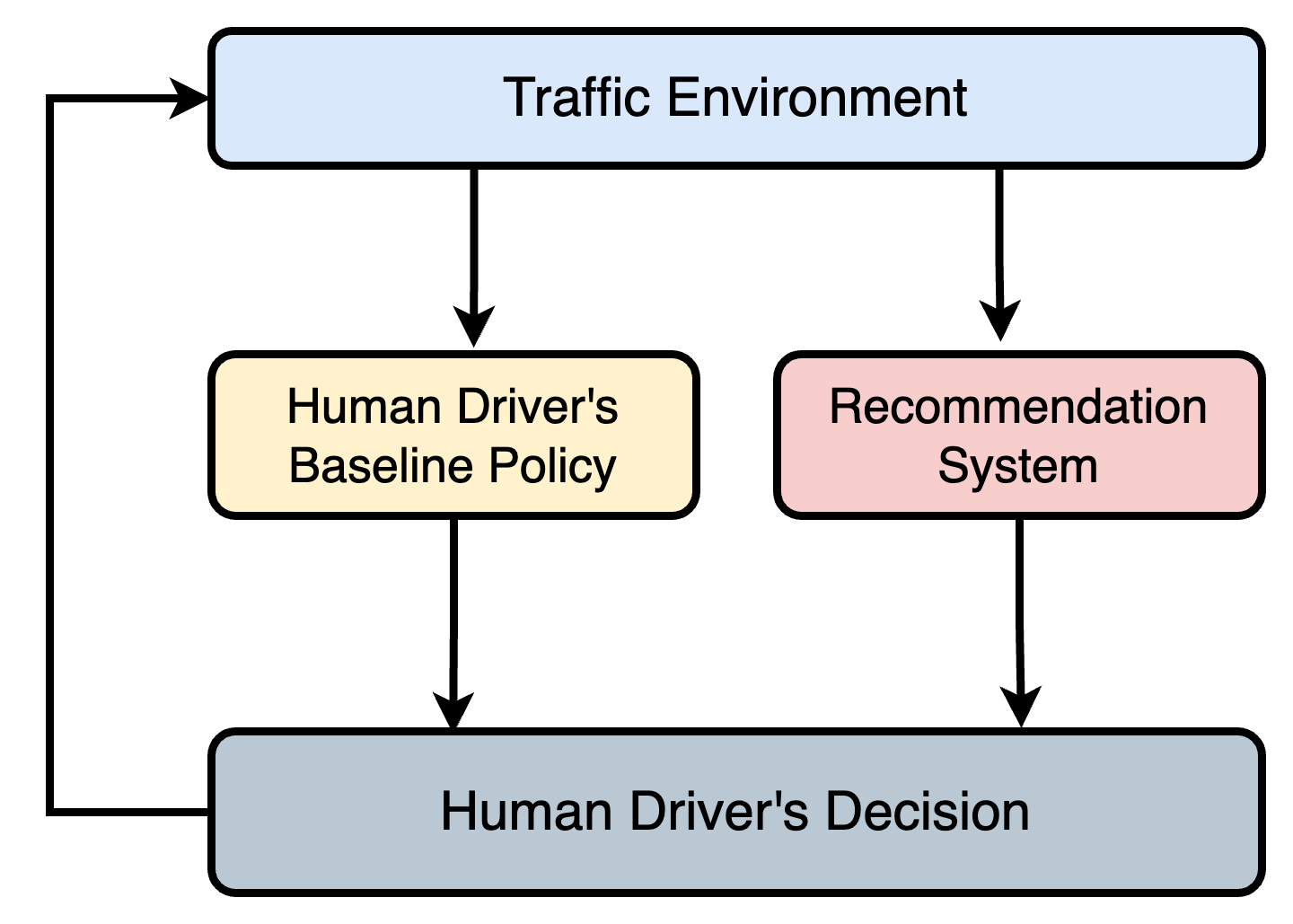}
    \caption{Process of decision-making.}
    \label{fig:system}
\end{figure}

In this section, we propose a solution approach that integrates the adherence-aware DQN agent into the system. We consider the lane-changing decision procedure as illustrated in Fig. \ref{fig:system}.

\subsection{Adherence-aware Q learning Approach}
To learn the human driver's adherence level to the recommendation on lane changing problem, we adopt the adherence-ware Q learning approach proposed by Faros \textit{et al.} \cite{faros_q-learning_2023}. In contrast to the conventional Q learning algorithm, adherence-aware Q learning is designed to optimize the recommendation policy, assuming the human drivers partially comply with the recommendation action, with adherence level $\theta$. An unbiased estimator is constructed to estimate the unknown $\theta$, which accounts for the action of the human driver in this problem context. A sequence of the independent and identical distributed random variables, $\{Y_t\}$, where $t\in\mathbb{N_+}$, denotes whether the human driver complies with the recommendation action at a certain time $t$. Specifically, $Y_t=1$ when $u(t)=g^r(x(t))$, and $Y_t=0$ when $u_t = g^b(x(t))$. To estimate the $\theta$ for any number of $n\in \mathbb{N}$, we need to collect a sample $w_t$ containing observations $Y_0, Y_1, ..., Y_n$ and use it to approximate $\theta$ as $\hat{\theta} = w_t/n$. Throughout the learning process, we can update the estimated $\theta$ at step $t+1$ by the following updating rule: 
\begin{equation}
    \theta_{t+1} = \frac{\theta_t\cdot n + \mathbb{I}(Y_t=1)}{n+1}, 
\end{equation}
where $\mathbb{I}(\cdot)$ is the indicator function of $Y_t$.
With sufficiently large $n$, it is known that the update rule will converge to the true $\theta$ value from the unbiased estimator \cite{lehmann2006theory}, i.e.
\begin{equation}
    \mathbb{E}[\theta_{t+1}] = \hat{\theta}.
\end{equation}
The adherence-aware Q learning framework updates the value of the Q function $Q(x,u)$ following the procedure in the standard Q learning algorithm, with an updating rule involving $\hat{\theta}$:
\begin{align}\label{eq:Q updating}
\begin{split}  
    Q(x,u) \xleftarrow{} &Q(x,u) + \alpha\cdot\bigg{\{}\hat{\theta}\cdot\big{[}r(x,u) \\
    & + \lambda\cdot\max_{u'^{\mathrm{r}}}{Q(x', u'^{\mathrm{r}})}\big{]}
    +  (1-\hat{\theta})\cdot \big{[}r(x,u) \\
    &+ \lambda\cdot Q(x', u'^{\mathrm{b}})\big{]}  - Q(x,u)
    \bigg{\}},
\end{split}
\end{align}
where $x', u'$ are the state and action of the next time-step, with a learning rate $\alpha$ and a discount factor $\lambda$. In the work of Faros et al., the convergence for the adherence-aware algorithm to a certain optimal Q function is proved \cite{faros_q-learning_2023}. In this approach setting, the adherence level $\theta$ and the transition probabilities are unknown to the algorithm. The baseline policy is explicitly known and arbitrarily defined.

\subsection{Deep Q Network}

In conventional RL settings, deriving an optimal policy becomes computationally challenging when dealing with continuous state space. Classic Q learning approaches relying on the Q table to represent state-action values explicitly are infeasible to obtain the optimal policy. Since the state space includes speed and position coordinates that are continuous, a DQN framework is used to combine the adherence-aware Q value updating rules with a deep neural network. Instead of comparing all transition cases across Q values to find the optimal in each intermediate step, the DQN randomly draws a mini-batch of transactions to include in the Q function update \cite{mnih_human-level_2015}. The learning process is driven by minimizing a loss function measuring the means square error between the current Q value and target Q value derived from adherence-aware Q updating rule \eqref{eq:Q updating}. The loss function of given time $t$ is defined as
\begin{equation}
    L_t(\beta_Q) = \mathbb{E}_{x,u,r,u'}\big{[}(y_t-Q(x,u;\beta_Q))^2 \big{]},
\end{equation}
where $\beta_Q$ is the parameter of the neural network, and $y_t$ is the target Q value derived from the adherence-aware Q learning updating rule:
\begin{align}
\begin{split}
        y_t = \hat{\theta}\cdot& \big{[} r_t(x,u) + \lambda\cdot\max_{u'^r}{Q(x',u'^r; \beta_Q)} \big{]}\\
        &+ (1-\hat{\theta})\cdot\big{[} r_t(x,u) + \lambda\cdot Q(x', u'^b; \beta_Q) \big{]}.
\end{split}
\end{align}

The above formulation allows the DQN agent to learn the driver's adherence level similarly to the standard adherence-aware Q learning in \cite{faros_q-learning_2023} while updating learning loss.

\subsection{Implementation}
We use the CARLA simulation environment to test and evaluate the adherence-aware DQN algorithm. CARLA is an open-source driving simulator for transportation research, especially in autonomous driving fields \cite{carla}. It has the capability to provide complex and dynamic mixed-traffic environments containing realistic traffic settings.
Moreover, CARLA provides APIs, which can be utilized to implement and connect external learning algorithms, as well as configure and modify both upper and lower-level controllers. Additionally, CARLA's integrated autonomous driving system allows vehicles to plan trajectories and control themselves without further implementation, which simulates the driver's common driving behaviors in such a dynamic driving environment.

\begin{figure}[th]
    \centering
    \includegraphics[width=\linewidth]{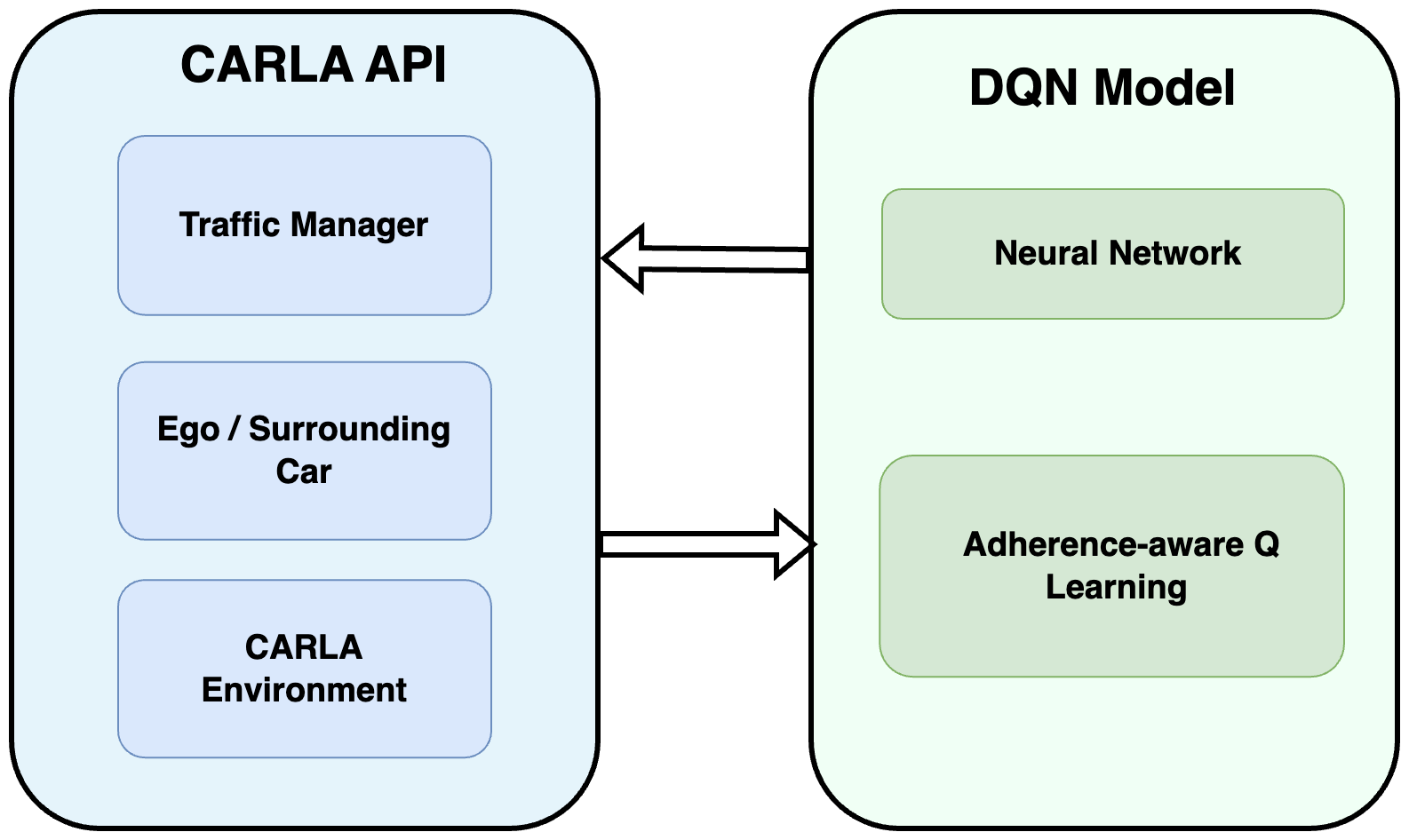}
    \caption{Software structure.}
    \label{fig:software}
\end{figure}

\section{Simulation}

\subsection{Simulation Environment}

 In this study, the entire simulation process is performed in the CARLA driving environment, including the learning process and evaluation. The physical laws and internal upper-level controllers in CARLA allow all the vehicles, including both ego and surrounding vehicles, to drive smoothly.
 % while lane changing recommendations are not provided.
 The overall software structure of modeling the simulation is shown in Fig. \ref{fig:software}.
 % The simulation is performed in CARLA's additional map, \textit{Town06}, which contains a long-distance highway with multiple lanes, entrance and exit on both sides, allowing enough space for executing multiple lane changing maneuvers. 
 % The map is loaded and rendered at the moment of starting the CARLA simulator when connecting to the server. 

Python APIs are utilized to connect the implemented adherence-aware DQN agent to CARLA and exchange data in real time when running the experiments. 
Since there are no lower-level controllers in the modeling framwork, we use CARLA's autopilot feature to represent the movement of a human driving a vehicle.
We modify the maximum speed of CARLA's autopilot module and lane-changing controller using the traffic manager API \cite{carla_tm}. This allows us to define the desired speed limit of all vehicles in autopilot mode and to turn autonomous lane-changing behavior on or off in autopilot mode. Hence, it simulates the cases when the ego vehicle only executes actions when human drivers comply with recommendations or their baseline actions while not complying.
 % These features can be applied seamlessly to the simulation setting of this study. 
 % We control the speed limit lower than the highway setting to simulate a more realistic case when multiple vehicles form congestion and slow down the traffic. 
 % \textcolor{blue}{Additionally, we force to disable the autonomous lane changing when autopilot is on to ensure that the ego car only executes lane-changing actions recommended by our algorithm.} 
 
Moreover, python APIs provide the ability for real-time data exchange, which makes it feasible to obtain online state input to the DQN model, as well as other related information. We define an additional software class containing functions to pass the state to the training loop connecting to the DQN, and safety modules to detect collision and distance to surrounding vehicles when switching lanes.

\subsection{Scenario}

\begin{figure}[h]
    \centering
    \includegraphics[width=0.8\linewidth, height=7.5cm]{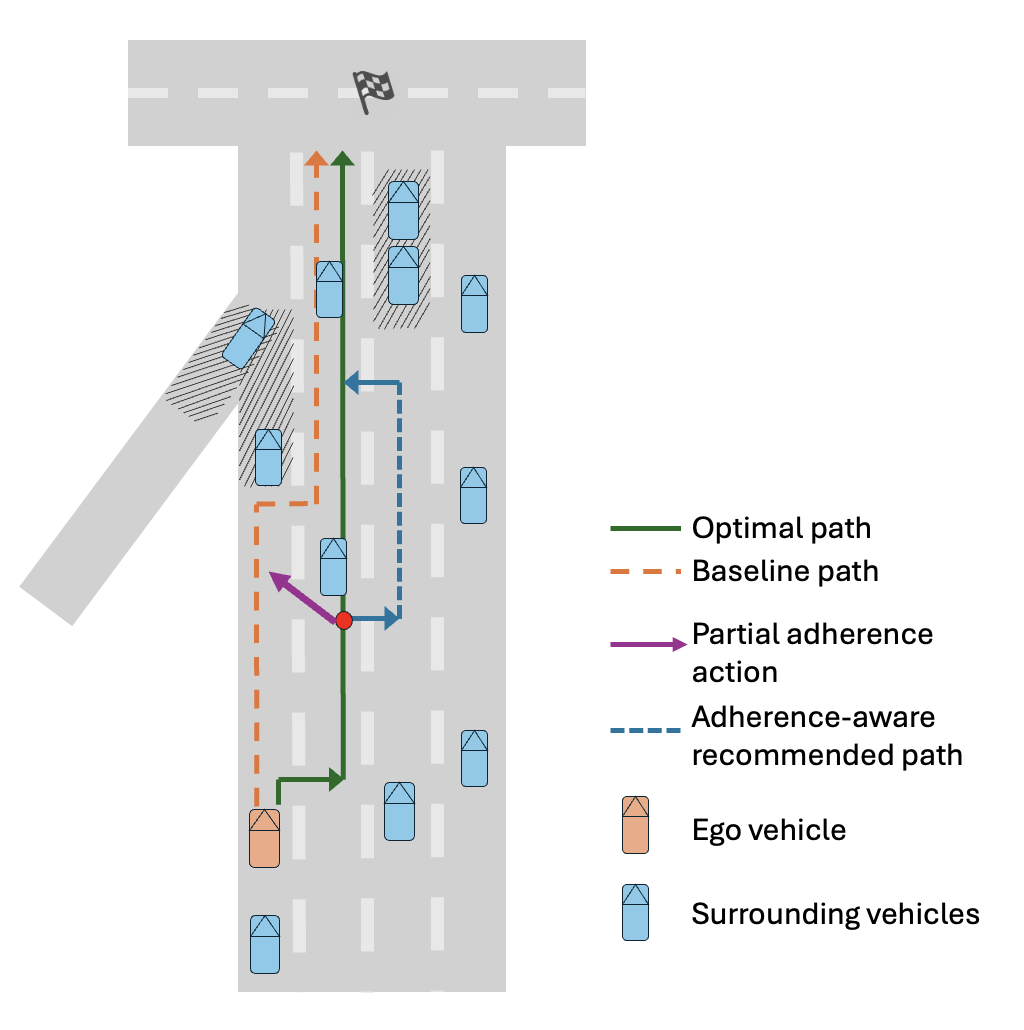}
    \caption{Simulation scenario where partial adherence leads to slower traffic, with different types of paths.}
    \label{fig:scenario}
\end{figure}

A city freeway case is considered for the simulation. In CARLA's map, \textit{Town 06}, a straight 4-lane freeway fragment is utilized. The length of the freeway is 300 meters, and the width of each lane is 3.5 meters. The default speed limit is 55 km/h. The freeway has a one-lane entrance from the left and ends with an intersection with only straight and right directions available, and the right two lanes are right-turning only lanes. Hence, vehicles are expected to perform mandatory lane changing before driving into the solid line area \cite{mandatory-lc}. If there are more vehicles in this area with different desired destinations, it is more likely to cause congestion or slow down traffic.
% From this road map, congestion can be formed due to the slow merging of vehicles from the entrance on the left side. 
The setting is roughly illustrated in Fig \ref{fig:scenario}. 

The ego vehicle is initialized at the leftmost lane in each episode, with multiple vehicles generated at certain positions and vehicles merging into the freeway sententially from the entrance. We also generate several vehicles randomly in each episode. During the training, the state of all vehicles can be obtained from CARLA using APIs in real-time. We also add the lateral coordinate in the computation to align with CARLA's data structures.

During ego vehicle's driving, we automatically capture the five surrounding vehicles denoted by $\mathcal{V}$, regardless of the total number of vehicles in the scenario. Moreover, suppose the ego vehicle is running on the leftmost or rightmost lane. In that case, we increase the ego car's position by a large distance to represent the position of surrounding vehicles that could have been on the left and right lane, respectively, and set the speed of these vehicles to be equal to the ego vehicle.

We set the true $\theta$ to be $0.5$ for the DQN agent to estimate and learn the partial adherence level. Under this setting, the human driver executes the recommended action with a probability of 50\% and executes the baseline action for the rest. We define the baseline action such that the vehicle follows the preceding vehicle on the current lane but only changes the lane to the available lane when the speed is slower than a threshold to represent running into congestion. If both directions are available to change lanes in the baseline action, there is an equal probability of changing to either direction.

We train the DQN agent online with the CARLA environment running. The agent is trained through 3000 episodes.
% with the training time of approximately half a minute for each episode. 
% One time step $\Delta t$ is $0.1s$. 
% During the lane changing process, the data are excluded from learning since the vehicle's behavior can cause confusion to the learning algorithm. 
The neural network to estimate the target Q value is designed to contain one hidden layer with 128 neurons. 
% A mini batch with is utilized to store the information for transitions, including states, recommended action, baseline action, and actual action. 
We use a decayed epsilon strategy while optimal action searching to balance the trade-off between exploration and exploitation, with a decay factor of 0.995. We use the ReLU function as the activation function. The neural network is updated after each training episode, with loss calculation and Q value updating. Details of the neural network parameters are listed in Table. \ref{tab:dqn-para}.

\begin{table}[t]
\centering
\caption{DQN parameters}
\begin{tabular}{l|c}
\hline
\textbf{Parameters} & \textbf{Values}\\
\hline
\textbf{Adherence Level}    & 0.5  \\
\textbf{Learning Rate}      & 1E-6   \\
\textbf{Discount Factor}           & 0.95  \\
\textbf{Minimum Epsilon}    & 0.001  \\
\textbf{Maximum Epsilon}    & 1 \\
\textbf{Epsilon Decay}      & 0.995  \\
% \textbf{Input Neuron Number} & 24 \\
\textbf{Hidden Layer Neuron Number} & 128 \\
% \textbf{2nd Hidden Layer Neuron Number} & 128 \\
\textbf{Batch Size}         & 32  \\
\textbf{Number of Episodes} & 3000\\
\hline
\end{tabular}
\label{tab:dqn-para}
\end{table}

\subsection{Results}

\begin{figure}[t]
    \centering
    \includegraphics[width=\linewidth]{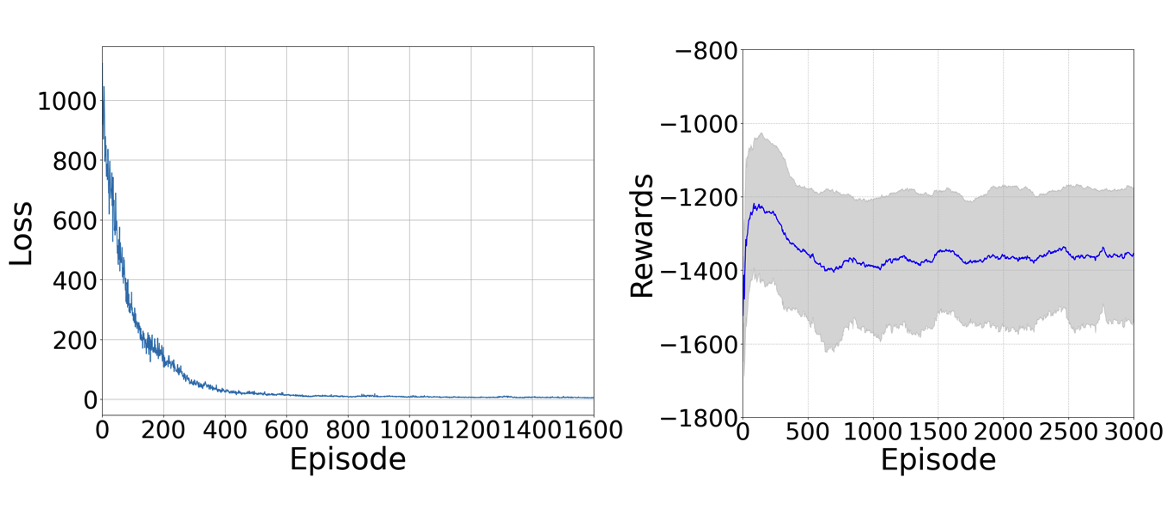}
    \caption{Training loss and smoothed cumulated reward with shaded fluctuation.}
    \label{fig:converge}
\end{figure}

\begin{figure*}
    \centering
    \includegraphics[width=0.75\linewidth, height=6.1cm]{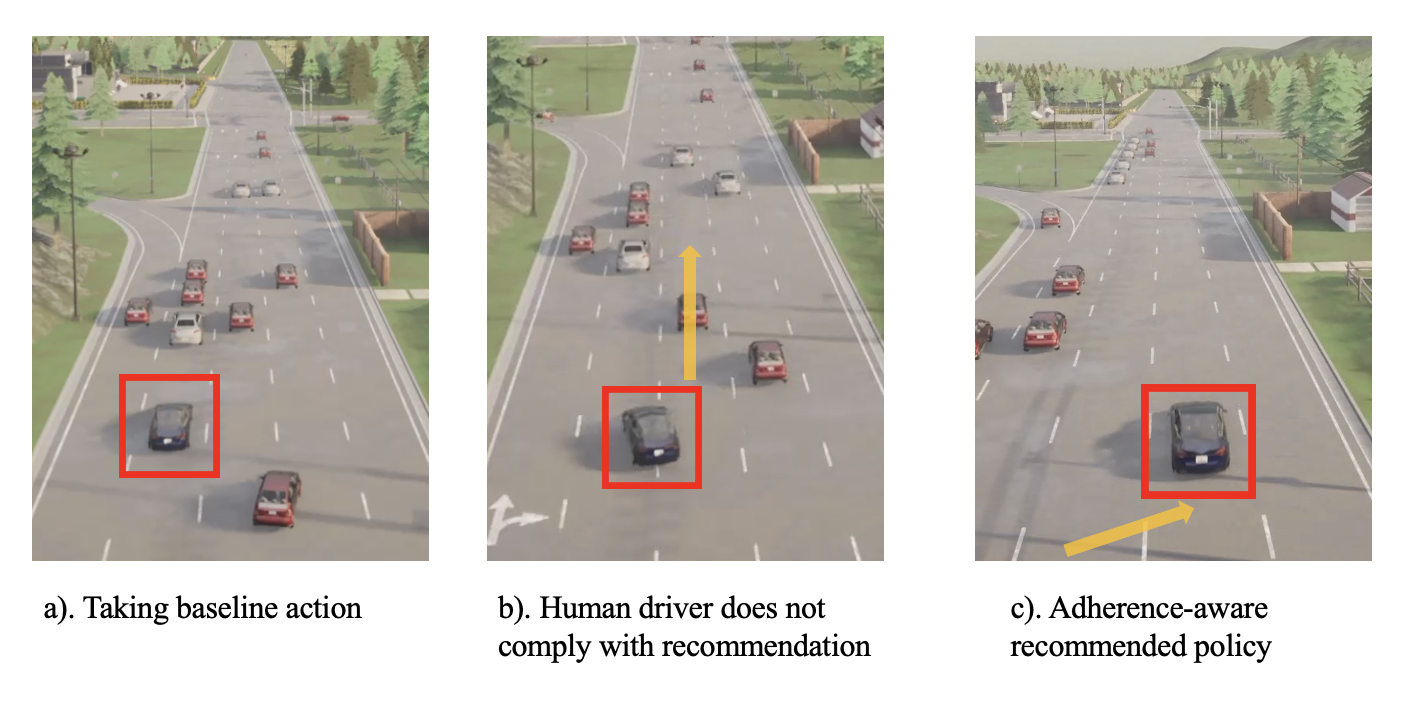}
    \caption{Lane changing decisions.}
    \label{fig:carla}
\end{figure*}

The DQN loss values and the cumulative rewards per episode converge to a stable value through the training as shown in Fig. \ref{fig:converge}. We also observe a decreased trend in the loss value and an overall increment of the reward.

We compare the adherence-aware DQN performance, the baseline policy, and a regular DQN agent regarding metrics in Table \ref{tab:performance_comparison}. Note that the human driver's partial compliance becomes a random disturbance in the regular DQN training process. From the results, following the optimal adherence-aware policy reduces the travel time by $10.76\%$ compared to the baseline. Compared to the regular RL approach on the total rewards, the adherence-aware RL increases the reward by $31.33\%$. One particular criterion to notice is the safety cost. Since following the baseline action would result in minimum safety risk due to lane-changing actions, and other reward terms (e.g., speed reward) take the lead in reward computation, it can be explained that taking baseline action achieves the optimal safety cost. Overall, The simulation result shows an improvement in travel efficiency when the human driver conducts recommended action half of the time, using the adherence-aware DQN. 

Additionally, we demonstrate the ego vehicle's different behavior in several cases in CARLA, as shown in Fig. \ref{fig:carla}. The ego vehicle is highlighted in the red box, and the yellow arrows represent the recommendation. Fig \ref{fig:carla}. a) shows the baseline action, resulting in running into traffic congestion. In Fig. \ref{fig:carla}. b), the recommendation suggests keeping the current lane. However, with the human driver's partial adherence, the ego vehicle essentially takes a left lane change and will result in congestion. Fig \ref{fig:carla}. c) shows the adherence-aware algorithm considers the human driver's possible lane changes opposing the recommendation. To avoid human drivers entering the congested lane, our algorithm suggests moving to the right lane so that the ego cars can stay away from the congested lane.

\begin{table}[H]
\centering
\caption{Performance comparison}
\begin{tabular}{l|ccc}
\hline
\textbf{Metric} & \textbf{Baseline} & \multicolumn{1}{p{0.4cm}}{\centering\textbf{Regular} \\ \centering\textbf{RL}} & \multicolumn{1}{p{2.15cm}}{\centering \textbf{Adherence-aware} \\ \textbf{RL}} \\
\hline
\textbf{Average Speed (km/h)}      & 23.47    & 29.56    & 31.02    \\
\textbf{Travel Time (s)}           & 38.83    & 35.05    & 34.65    \\
\textbf{Cumulative Reward (shifted)}    & 265.97 & 600.79 & 874.9 \\
\textbf{Speed Reward (shifted)}    & 476.01 & 850.03 & 1147.92  \\
\textbf{Unnecessary-changing Cost} & -29.50    & -22.69    & -3.29     \\
\textbf{Safety Cost}               & -150.78   & -198.62   & -179.17   \\
\textbf{Missing-changing Cost}     & -29.76    & -27.95    & -23.09    \\
% \textbf{Average Reward per time}   & -24.72   & -21.44   & -16.83   \\
\hline
\end{tabular}
\label{tab:performance_comparison}
\end{table}

% \newpage

\section{Concluding Remarks} \label{sec:conclusion}

In this paper, we introduced the adherence-aware reinforcement learning method for solving the lane-changing decision problem. We modeled the problem using an MDP that incorporated the partial adherence level from human drivers. The proposed algorithm estimated the adherence level and computed the optimal policy in real time within an extended DQN framework. To implement the solution approach, we trained the model and simulated the results in the CARLA environment. The simulation demonstrated that the adherence-aware DQN was capable of learning the partial adherence level and providing a suboptimal solution regarding human drivers’ deviations from the recommendations. The adherence-aware RL approach outperforms both the policy derived from a regular RL method and the baseline policy from the perspective of travel efficiency, given the partial compliance level of human drivers. 
Potential extensions of this work include modeling the problem with a dynamic compliance level throughout driving and different types of baseline actions. Future research should also involve improvements to the adherence-aware Q algorithm or training the DQN in more complex scenarios with a more advanced network architecture.

\bibliographystyle{IEEEtran}
\bibliography{CPS,IDS}

\end{document}